\documentclass[conference]{IEEEtran}
\IEEEoverridecommandlockouts
\usepackage{cite}
\usepackage{amsmath,amssymb,amsfonts}

\usepackage{graphicx}
\usepackage{textcomp}
\usepackage{xcolor}

\usepackage{preamble}

\def\BibTeX{{\rm B\kern-.05em{\sc i\kern-.025em b}\kern-.08em
    T\kern-.1667em\lower.7ex\hbox{E}\kern-.125emX}}

\begin{document}

    \title{A Monte Carlo Approach to Closing the Reality Gap\thanks{Lyons is partially supported by grant DL-47359-15016 from Bloomberg 
The views and conclusions contained herein are those of the authors and should not be interpreted as necessarily representing the official policies or endorsements, either expressed or implied, of the U.S. Department of Defense or the U.S. Government.}
}

\author{\IEEEauthorblockN{Damian Lyons}
\IEEEauthorblockA{
\textit{Fordham University}\\
Bronx, New York\\
{\footnotesize dlyons@fordham.edu}}
\and
\IEEEauthorblockN{James Finocchiaro}
\IEEEauthorblockA{
\textit{United States Military Academy}\\
West Point, New York\\
{\footnotesize james.finocchiaro@westpoint.edu}}
\and
\IEEEauthorblockN{Michael Novitzky}
\IEEEauthorblockA{
\textit{United States Military Academy}\\
West Point, New York\\
{\footnotesize michael.novitzky@westpoint.edu}}
\and
\IEEEauthorblockN{Christopher Korpela}
\IEEEauthorblockA{
\textit{United States Military Academy}\\
West Point, New York\\
{\footnotesize christopher.korpela@westpoint.edu}}
}

    \maketitle

    \begin{abstract}

We propose a novel approach to the 'reality gap' problem, i.e., modifying a robot simulation so that its performance becomes more similar to observed real world phenomena. This problem arises whether the simulation is being used by human designers or in an automated policy development mechanism. We expect that the  program/policy is  developed  using  simulation,  and  subsequently deployed  on  a  real system.  We further assume  that  the  program includes a monitor procedure with scalar output to determine when it is achieving its performance objectives.
The proposed approach collects simulation and real world observations and builds conditional probability functions. These are used to generate paired roll-outs to identify points of divergence in behavior. These are used to generate {\it state-space kernels} that coerce the simulation into behaving more like observed reality. 

The method was evaluated using ROS/Gazebo for simulation and a heavily modified Traaxas platform in outdoor deployment. The results support not just that the kernel approach can force the simulation to behave more like reality, but that the modification is such that an improved control policy tested in the modified simulation also performs better in the real world.

\end{abstract}
    \section{Introduction}\label{sec:introduction}

Simulation tools are widely used in robot program development, whether the program is built by hand  or using machine learning.
Simulation allows a robot programmer to eliminate most obvious program flaws without damaging hardware. The availability of physics engines \cite{Reckhous_2010} has produced simulations that can more accurately model physical behavior has made it more attractive to use simulation in conjunction with machine learning techniques\cite{Peng_2018}\cite{moeckel_2013} to develop robot programs.  However, a robot program validated with simulation, when operating in a real, unstructured environment may come across phenomena that its designers just did not know to include in the simulation, even though the phenomenon could in fact be simulated if it were a-priori known to be relevant. Examples of this kind of  simulation {\it `reality gap'} include inaccurate robot joint parameters,  surface friction, object masses, sizes and locations. 

This paper addresses closing the {\it reality gap} for simulations used to develop robot programs. We will assume that the simulation is black-box and we present a wrapper framework that can be used to coerce  simulation behavior to more closely resemble collected experience. 
The simulation has configuration parameters $\phi$. If deployment of the control software ($\pi$)
results in failure as determined by the performance monitor, then our overall objective is for $\phi$ to be updated by the deployment experience and $\pi$ redeveloped to handle that experience. 
We expect that this virtuous cycle of simulation,  deployment and improvement to iterate.
The simulation will be modeled as a transition function $T_{\phi}(s' | s , a)$ where $a$ is the action to be carried out, $s$ is the current sensor data from the simulation, $s'$ is the resulting sensor data, and $\phi$ is a setting of the configuration parameters for the simulation. Real world experience will also be modeled as a transition function $T_r(s' | s , a)$. Two important novel aspects of our work are 1) a domain independent proposal for $\phi$ (as opposed to the more domain specific examples of domain randomization, e.g., \cite{Tobin_2017}\cite{Dehban_2019}) and 2) a Monte Carlo method to modify $\phi$ based on a direct comparison of estimated transitions functions (as opposed to the comparison of observations).

\begin{figure}
\begin{center}
 \includegraphics[width=3.5truein]{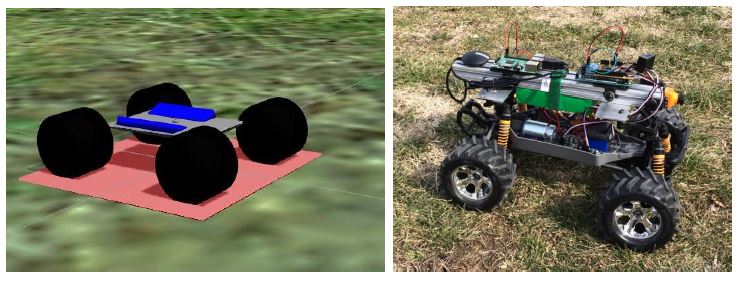}
 \caption{Simulated and actual platforms used in the field trials.}\label{fig:platforms}
 \end{center}
 \vspace{-0.35in}
\end{figure}

The remainder of the paper is laid out as follows. Section II presents a review of related work. Section III presents our method and architecture. 
Section IV presents a 1-D example to illustrate the approach. Section V describes the main experiment, using a ROS/Gazebo Ackermann vehicle simulation for design and a customized Traaxis platform for deployment, Fig. \ref{fig:platforms}. Section VI presents our results, showing that data collected during deployment can be used to make the ROS/Gazebo simulation behave more realistically, iteratively narrowing the reality gap. 
Conclusions and next steps are presented in Section VI.

    \section{Related Work}\label{sec:related_work}

Addressing the problem of making a robot simulation produce more realistic behavior does not just concern improved physics engine and rendering technology, but also concerns handling uncertainty about which specific environment the robot will encounter. Benjamin \cite{Benjamin_2013} and Lyons \cite{Lyons_2012} proposed an approach for cognitive robotics where a simulation is used to predict the environment state and to visually compare video with 
simulation predicted imagery. Differences are used to modify the simulation, changing object appearance, positions and velocities to match the observed video. A great deal of domain information is necessary to support this simulation update step. 

Researchers using learning techniques to train robot programs on simulation have developed several approaches to the reality gap. The {\it sim-to-real} approach addresses the issue of moving a policy trained in simulation to real hardware. Christian et al. \cite{christiano_2016} use a simulation to learn policies for a number of tasks and consider transferring the policy from simulation to real robot. They note that the simulation policy is generally correct in high-level gist but fails on some lower-level details. Yu \cite{yu_2019} proposes a two stage approach to policy training for bipedal locomotion: a presimulation step to ballpark the system identification for simulation, followed by a more accurate tuning at deployment. Permana \cite{permana_2019} trains a CNN on synthetic imagery for visual detection of ground casualties, and handles the sim-to-real issue by injecting noise, downsampling, segment removal and other changes to the simulation data.

Domain randomization is an approach with a similar effect, Peng et al. \cite{Peng_2018} use random modification of 95 simulation parameters to show that even low fidelity simulation can be used to train a robot to push a puck successfully over a wide range of real-world situations. 
Cheboter et al \cite{Chebotar_2019} interleave many simulation runs with much fewer real-world runs,  and modify the simulation configuration so that it simulates a range of situations more similar to observed experience. Our proposed approach is most like that of Chebotar. However, instead of using the domain randomization --- establishing a range of initial parameter values --- we follow an approach more similar to Benjamin and Lyons \cite{Lyons_2012} who modify simulation state during a simulation to close the reality gap. Our approach differs from sim-to-real approaches in putting the emphasis on improving the simulation so that the sim-to-real step becomes less onerous.

The proposed approach is a component of a self-healing approach to autonomous system software described in \cite{Lyons_2020}. The work describe here just concerns the reality gap
and could be used as part of any approach for which that is an issue.

    \section{Approach}\label{sec:approach}

The simulated and physical environments are modeled as Markov Decision Processes $M_{sim}=(S,A,T_{sim},R)$ and $M_{phy}= (S,A,T_{phy},R)$ that differ only in their transition function. The control software for the robot is a policy $\pi$ that can be applied to either MDP. 

The set of sensors available to the control software $SN={sn_0,sn_1,\ldots,Sn_n}$ and the value sets for each sensor are $SV={Sv_0,Sv_1,\ldots,Sv_n }$. The set of control outputs available to the control software are $AN={an_0,an_1,\ldots ,an_m}$ and the value sets for the control outputs are $AV={Av_0,Av_1,\ldots,Av_m}$. The actions available to the robot are any setting of the control outputs
\begin{equation}\label{eqn:a}
A = Av_0\times{}Av_1\times{}\ldots{}Av_m
\end{equation}
The set of states S is 
\begin{equation}\label{eqn:s}
S  =  Sv_0\times{}Sv_1\times{}\ldots{}Sv_n
\end{equation}
This state represents all the observations available to the control software. 
We will  assume that we always have access to all the sensors and not just the ones used for any task in progress. 

Optimization of the reward function, R, will be used as a measure of when the control software is achieving its performance objectives  on $M_{sim}$ or $M_{phy}$. The method for the calculation of R depends on how the control software was generated:
\begin{itemize}
\item Software designers build software guided by intuition \cite{Naur_1997} 
so that the software performs according to their apprehension of the specification. In that case,  we argue a software designer has the skills and intuition to instrument the software to measure whether the performance guarantee holds or not: That is, a performance monitor can be added to the control software to generate R. 
\item For code that is synthesized from a specification, R is must generated in a different way. The field of runtime verification addresses the issue of how to instrument a program to determine if it is behaving according to a specification \cite{Bartocci_2018}. 
\item Finally, for code that is learned, the reward structure from learning should  be used  to generate R during execution.
\end{itemize}
We will assume that the control software has been instrumented appropriately to deliver R when executed.

The MDP transition functions are defined
\begin{equation}\label{eqn:p}
T_{sim},T_{phy} : S\times S\times A \rightarrow [0,1]
\end{equation}
and are interpreted in their usual way as the conditional probability $T(s' | s, a)$ of transitioning from state $s$ to state $s'$ when action $a$ is carried out. These functions encapsulate any difference between what occurs when an action $a$ is carried out in the simulation in a state $s$ and when the same action $a$ is carried out in state $s$ in the real environment. 

\subsection{Kernel Generation}

Once the software designers have run whatever simulation and physical tests they decide are sufficient that the software can be deployed,  we collect data from multiple runs of this final version of the software through the simulation:
\begin{equation}\label{eqn:h}
H_{sim} = \{ ( s_i, a_i, r_i, s'_i ) : 0\leq i \leq I_{max}\}
\end{equation}
for all action $a$ taken in state $s$ then resulting in state $s'$ and reward $r$ in a run of the simulation. Using a frequentist approach, $T_{sim}$ is estimated from $H_{sim}$, and $T_{phy}$ from $H_{phy}.$ The state to state transition function $P$ that is the combination of the $T$ with $\pi$, the control software is:
\begin{equation}\label{eqn:p}
P(s' | s, a) = T(s'| s,a)\pi(a|s)
\end{equation}
Let $SR(s,a)$ be a description of what states can be transition to from $s$ with $P$:
\begin{equation}\label{eqn:sr}
SR(s,a) = \{ ( s' , P( s' | s ) ) : P(s'| s,a)> 0, a\in A \}
\end{equation}
Since a state only includes what is observable to the robot, there may be hidden dynamics in the real world that result in $SR_{phy}(s,a)$ differing from $SR_{sim}(s,a)$. Our first objective is to identify this difference. 
We define the policy roll-out from a state $s$ as:
\begin{equation}\label{eqn:roll}
\begin{array}{lcl}
Roll(s) & =  & (s', a, p) . Roll(s') , (s', p)  \in SR(s,a)
\end{array}
\end{equation}
A number of {\it paired roll-outs} are calculated $Roll_{sim}(s)$ and $Roll_{phy}(s)$ and compared up to the point at which they are considered to diverge.
The distance between two states is defined by a distance function $c(s,s')$:
\begin{equation}\label{eqn:c}
c(s,s') = \sum_i w_i | s_i-s'_i |
\end{equation}
 States differ if their distance is greater than a threshold $\epsilon_c$. If states differ, 
 and they have sufficiently different probabilities $p$ under each transition function, this is called {\it a roll-out divergence}. For each divergence detected, we construct a state-space kernel that we will use to modify the behavior of the simulation in that region of its state space so that it is more similar to observed behavior of the physical environment. 
 
 The key pieces of information in a kernel $k$ are:
 \begin{itemize}
 \item The {\it region of state space} in which it is active: $e_k=N(0,\sigma^2)$ is a univariate normal distribution of distance from  $s$, the state preceding the divergence,  where  $c_k(.)=c(s,.)$ is used to calculate the scalar distance between $s$ and any other state. For any state $s'$, $e_k \circ c_k(s')$ is  measure of how active kernel k is at $s'$.
 
 \item The {\it divergence probability distribution}: $\wp_k=(p_s, p_p)$, where $\delta=(s,a,p)$ is the common roll-out point just before divergence and  
 $\delta_s =(s_s, p_s, a)$
 in $Roll_{sim}(s_0)$ and $\delta_p=(s_p, p_p,a)$ in  $Roll_{phy}(s_0)$ are the points after divergence.

 \item The {\it kernel transfer function}: $f_k : S \times A \rightarrow S$. This mapping is composed by superposition of  linear functions $f_{k,a}~a\in A$ constructed by least-squares fit to the state and action data in the sequence starting with the state preceding divergence and including a small number of successor states. We will restrict the kernel to be active in a small region of the state space, and since the transfer function estimate is in a small region of the state space, we argue that this can function as a piecewise 
 linear approximation of a more complex function.
 \end{itemize}
Algorithm \ref{alg:kc} shows the algorithm for kernel construction.

\begin{algorithm}
\caption{Generate Kernels}\label{alg:kc}
\begin{algorithmic}
\Procedure  {GenKer}{$P_{sim},P_{phy}$}
    \State $S_0 = States(P_{sim}) \cap States(P_{phy})$
    \While{$i < N_{rollouts}$}
        \State $s_0 \sim  U(S_0)$ // sample uniform distrib
        \State $r_s = Roll_{sim}(s_0)$
        \State $r_p = Roll_{phy}(s_0)$
        \State $\delta, \delta_s, \delta_p = divergence(r_s,r_p)$
        \State // $\delta$ followed by $\delta_s,\delta_p$ in $r_s, r_p$ resp.
        \If{$\delta\neq null$} // construct kernel at $\delta$
            \State $e = N(s, \sigma^2)$ // $\delta=(s,a,p)$ eq.(\ref{eqn:sr})
            \State $d = normalized (p_s, p_p)$ // $\delta_s=(s_s,a_s,p_s)$, sim $\delta_p$
            \State $f_{a,s} = linearfit(\delta,\delta_p,\ldots)$ // short subseq $r_p$
            \State $AddKernel(e,d,f)$
        \EndIf
    \EndWhile
 \EndProcedure
\end{algorithmic}
\end{algorithm}

\subsection{Kernel Manager}
We consider the simulation as a black box process, but one over which we have some control. For example, in \cite{Chebotar_2019} 
a black-box simulation can configured by selecting initial parameters $\phi$ from a distribution $P_\phi$. They attempt to learn the distribution $P_\phi$ that best matches traces of real stored experience. We will also assume that we have access to a set of parameters $\phi$ for the black box simulation but more similar to those of \cite{Lyons_2012}which are checked by the simulation on each  time-step and not just initialization. 

\begin{figure}
\begin{center}
 \includegraphics[width=3in]{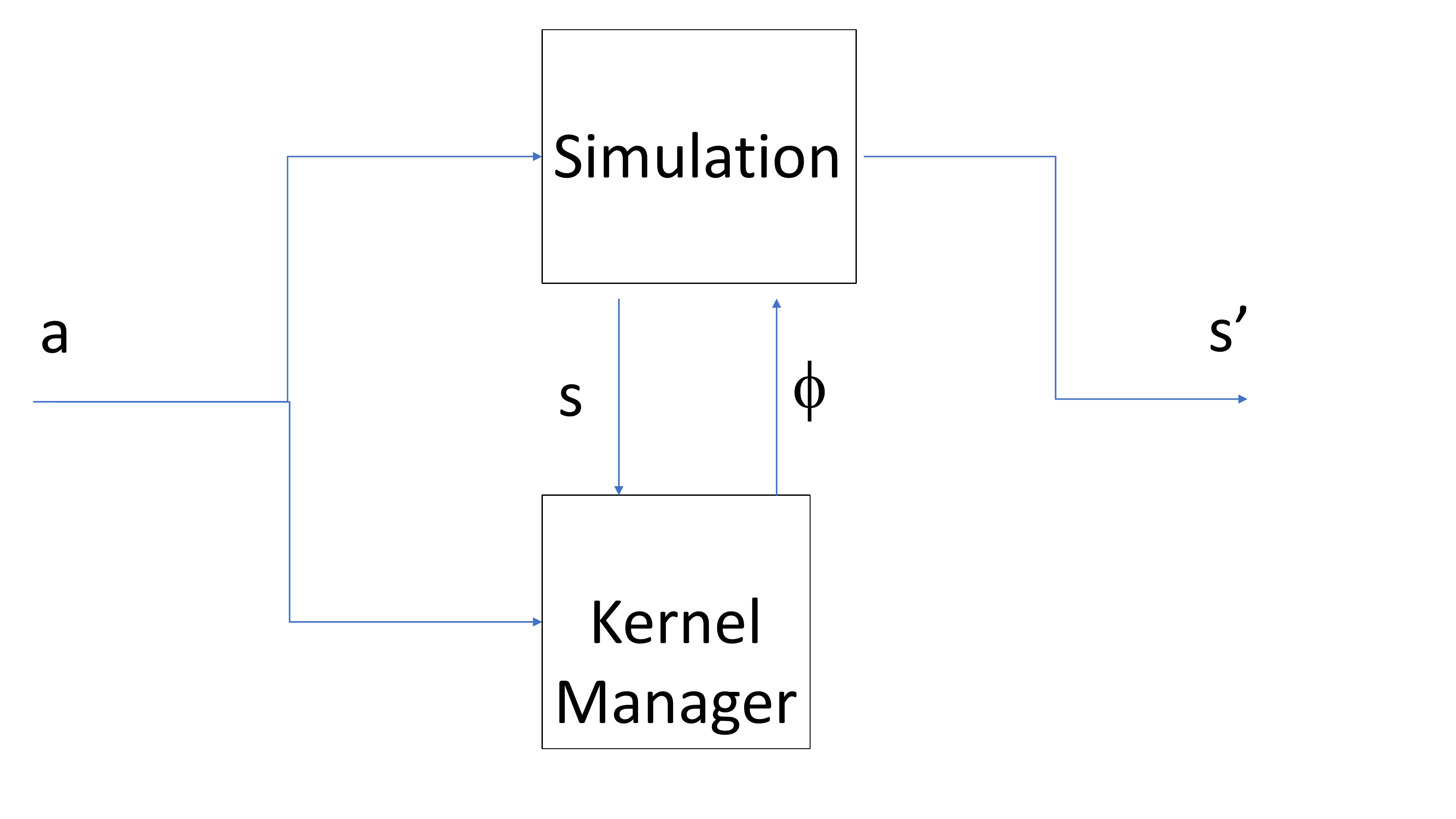}
 \vspace{-0.25in}
 \end{center}
 \caption{Architecture for kernel modified simulation}\label{fig:arch}
 \vspace{-0.2in}
\end{figure}

At each simulation time step, the control software sends its selected action $a$ to the simulation. While the internal state of the simulation is unknown, it generates 
the sensor output values, the $M_{sim}$ state $s$ and this is communicated to the {\it Kernel Manager}. The kernel manager checks first to see whether the state $s$ is within the spatial scope for any kernel $k$ by evaluating $e_k\circ c_k(s) > \epsilon_c$ (for small $\epsilon_c$) to see if that kernel is active.  
The kernel manager selects the first active kernel, if any are active, and uses it to modify the state of the simulation using the simulation parameters $\phi$ so that it returns $f_k(s,a)$. The kernel manager has no knowledge of the internal simulation state and dynamics. 

Although a kernel is only active in a small region of the state space, the effect of any change to the simulation state is potentially long lasting. This results in both a spatial and temporal generalization effect. A modification of an object location for example, will result in sensors picking up the revised location no matter the pose of the robot - a spatial generalization. Other simulation entities that subsequently interact with the object (e.g., collide) will do so at the new location - a temporal generalization. Both of these generalization leverage the simulation dynamics applied to new experience gained from execution of the control software in a real environment.

    \section{Example} \label{sec:ex1}

As a first example, consider a robot moving on a 1-D track back and forth between two waypoints - a simple model of an autonomous robot tasked with picking up and delivering supplies between several locations. The performance measure for the robot is that it achieves its waypoints within a specified accuracy and a specified time deadline. A reward of 10 is generated for on-time accurate arrival, 0 for any time before the deadline, and -10 for any time after the deadline. Although this would seem to indicate we are implying a reinforcement learning approach, we are not. The software could be  handwritten (as is ours Fig. \ref{fig:ex1code}) or any combination of both. All we insist is that the performance measure is in place (called for example by a hook function or timer) and the software (once ready for deployment) is validated in the simulation. 

\begin{figure}
\begin{verbatim}
def traveller(goal):
    global Epsilon # error threshold
    error,delta,velocity = 1, 0, 0
    while error>Epsilon:
        pos = roslisten('/robot/odom') #fake ROS topic
        delta = goal - pos 
        error = abs(delta)  
        vel=5*delta 
        rospublish(vel,'/robot/vel') # fake ROS topic
\end{verbatim}
\caption{Section of python control software for example 1}
\label{fig:ex1code}
\vspace{-0.2in}
\end{figure}

\begin{figure}
\begin{center}
 \vspace{-0.2in}
 \includegraphics[width=3in]{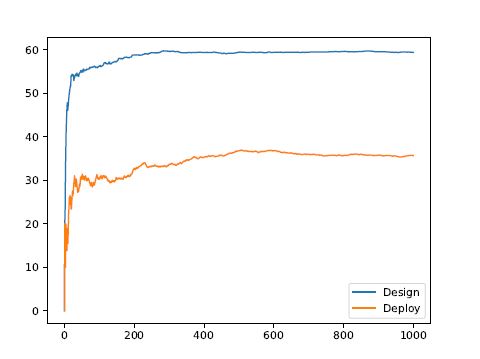} 
 \vspace{-0.1in}
 \end{center}
 \caption{ATR graphs for control software at design and deployment for 1-D example 
 } \label{fig:atr1}
\vspace{-0.1in}
\end{figure}

The simulation models the robot as a point. It accepts velocity as input each time step and generates a new position as output.
The position of the robot at the start of each mission is selected from a uniform distribution. The Average Total Reward (ATR), the sum of all rewards divided by number of missions, is used as a measure of success. Figure \ref{fig:atr1} shows the ATR graphs for the design and deployment of this example. The graph labeled ``Design'' is from the validating the software on the simulation. At this point in the design process, the software designers have all the information they consider necessary to have the software fulfill their understanding of the specification. Thus, we consider the ATR graph to be an objective  measure of what "success" should look like. 

The simulation can be configured so that a region of space becomes more difficult to traverse  - abstractly modeling an unexpected incline, rough terrain, mud, etc. The simulation also generates a binary terrain sensor value which becomes 1 iff the robot is within this region. To emulate our software encountering an unexpected environment, we configure the simulation to generate difficult regions randomly and at random points along the robot's traverse. When our control software (Fig. \ref{fig:ex1code})  encounters such a region it may fail to meet a deadline. The ATR graph in Fig. \ref{fig:atr1}, labeled ``Deploy'' shows the performance of the control software in this modified simulation (taking the place of a real deployment). It is lower than the design ATR. We consider this as an empirical indication that the software is not meeting its design specification; however, we don't know yet what is different about the environment at deployment time.

\begin{figure}
\begin{verbatim}
[ (15.0, 0.0),    # mean for e_k
  [0.51,0.48],    # distribution (p_s, p_p)
  [0.33, 0.33]  ] # f_k,v linear trans coeff
\end{verbatim}
\caption{Example of kernel found for 1-D example}\label{fig:ker}
\vspace{-0.2in}
\end{figure}

The control software is instructed to log H (eq.(\ref{eqn:h})). The state in this example includes the robot optometry and the terrain sensor "topics". The action value is the robot velocity "topic". The reward is the value from the performance metric. $H_{sim}$ and $H_{phy}$ were collected for the 1000 missions graphed in Fig. \ref{fig:atr1}.
The kernel generation algorithm was run on this data and 23 unique kernels were constructed (omitting duplicates and candidate divergences that did not have sufficient information to estimate the transfer function). One such kernel is shown in Fig.\ref{fig:ker}. 

\begin{figure}
\begin{center}
 \vspace{-0.3in}
 \includegraphics[width=3in]{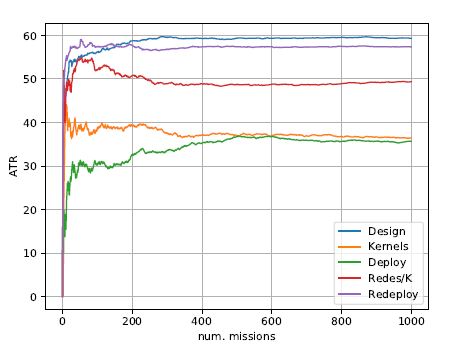} 
 \vspace{-0.15in}
 \end{center}
 \caption{ATR graphs for control software redesign in 1-D example}\label{fig:atr2}
  \vspace{-0.2in}
\end{figure}

The simulation configuration parameters allowed the position of the robot to be reset and the terrain sensor to be set or cleared. The control software is rerun through the simulation receiving configuration commands from the Kernel Manager (Fig. \ref{fig:arch}), generating the ATR graph labeled ``Kernels'' in Fig. \ref{fig:atr2}. It follows the ``Deploy'' graph after an initial period in which it is not as poorly performing as real deployment, but  much worse than the "Design" curve. A SARSA reinforcement learning algorithm was used to select velocity values given the sensor data and trained on the kernel modified simulation. The redesigned software shows improved performance of approx. 25\% , as shown by the "Redes/K" graph in Fig. \ref{fig:atr2}. Finally, the redesigned software
was run again in the ``Deploy'' situation - that is, in the simulation without any Kernel manager input, but with the difficult terrain feature enabled with the same settings as before. The graph in Fig. \ref{fig:atr2} labeled ``Redeploy'' shows the ATR for this case. It is higher than the original ``Deploy'' graph and close, but not identical to, to the original ``Design'' graph. The simulation, enhanced by the kernels extracted from prior physical experience, is now a better reflection of the physical environment.

    \section{Field Test}\label{sec:field}
To evaluate the proposed approach, we prototyped mission software for a three GPS waypoint mission on an mobile robot with Ackermann steering driving on flat terrain.  The mission software was then deployed to a physical robot for traversing the GPS waypoints in an open field, with the unexpected situation that the robot sometimes encountered a steep ramp between two of the waypoints. The short deployment trial experience was used to generate state-space kernels, which were used to reduce the reality gap for the simulation. The experimental procedure and mechanism is described in more detail below, and the results are presented in the next section.

\begin{figure}
\begin{center}
 \includegraphics[width=3.5truein]{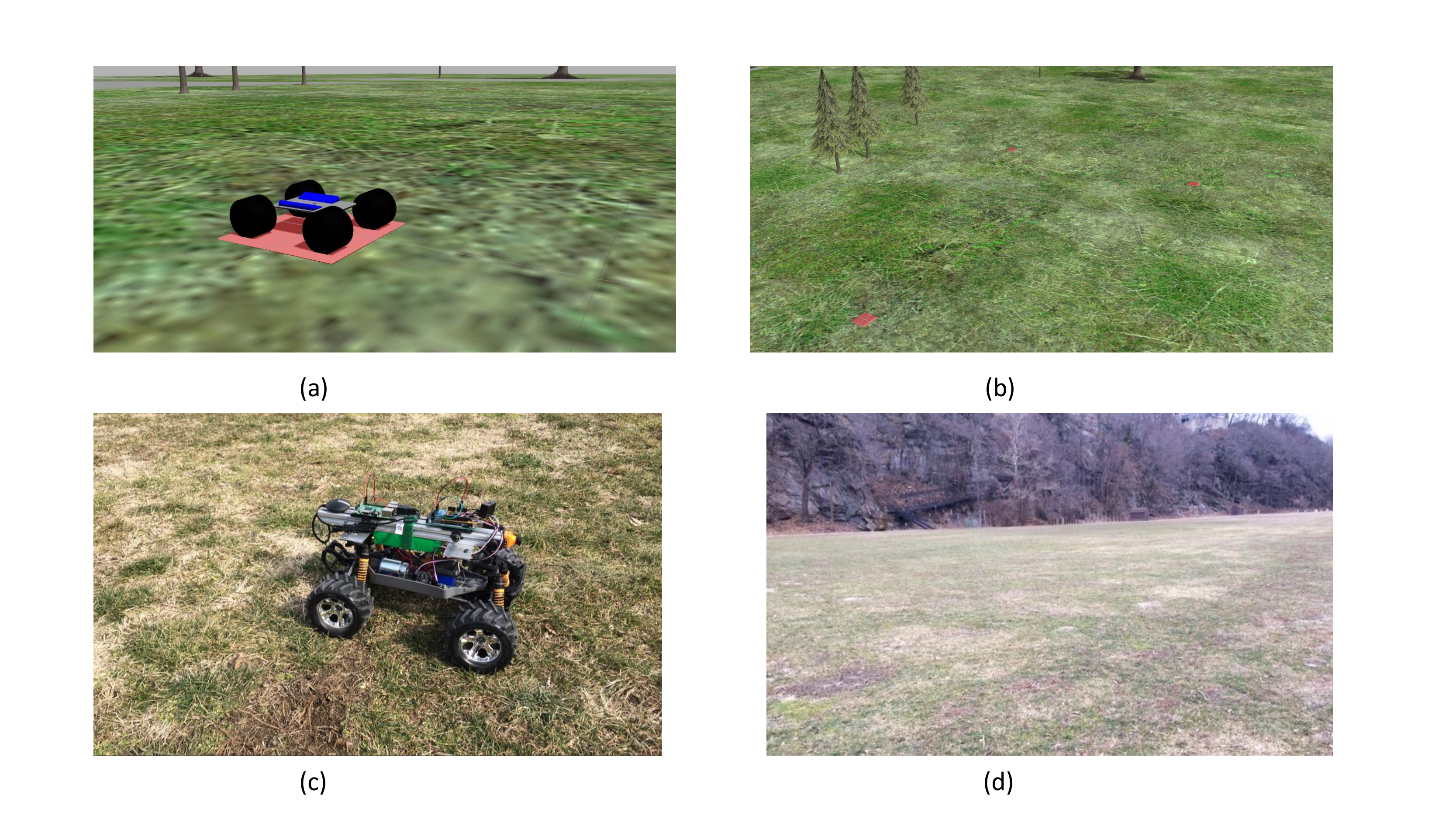}
 \vspace{-0.2in}
 \end{center}
 \caption{Simulation robot (a) and course (b), and real robot (c) and course (d)}\label{fig:field}
\vspace{-0.2in}
\end{figure}

\subsection{Design and Simulation}
Design and simulation work was completed using ROS Indigo  on a Dell Latitude 3460 laptop. The mission software was written in Python 2.7 and simulated using Gazebo with the open source Ackermann\_vehicle model\footnote{github.com/trainman419/ackermann\_vehicle-1}, Fig. \ref{fig:field}(a). An flat field was constructed using open source Gazebo models with three marked waypoints in world coordinates: (0,0), (5m,15m) and (-5m,15m), Fig. \ref{fig:field}(b). The mission control software used a saturated spring model to generate velocities towards waypoints, and a bicycle pursuit model to generate the steering and speed for the vehicle, running at 10 Hz. GPS data was simulated by providing Gazebo model position information at a rate of 1Hz, and vehicle orientation was calculated from GPS velocity. Vehicle pitch sensing was also simulated from model information.

Ackermann kinematics were used to interpolate position between GPS samples. A small amount of uniformly distributed noise was added to the steering and speed signals to simulate moving on bumpy grass. The performance requirement was that the vehicle complete the course within a time deadline of 39 seconds (empirically established for the experiment) and with an accuracy of 2m of the final waypoint. The software was instrumented to collect all sensor, control and reward signals at a rate of 0.5 Hz. 

\subsection{Description of the Field Trials}
The field trials were performed at the River Courts located at the United States Military Academy (USMA) in West Point, NY.  The Robotics Research Center (RRC) at USMA uses the River Courts to perform robotics testing for aerial and ground robots.  To facilitate testing, they have a trailer designed to be used/moved to remote locations for field trials and provide researchers with electricity, wifi, and climate control.  

The Traxxas Stampede 4x4 VXL, a four-wheel drive remote controlled vehicle, was used as the base of our platform for conducting experiments. We modified the standard platform physically by removing the truck plastic body and then changing out the shocks with stiffer springs to allow the platform to carry more weight. The standard platform weighs 2.54 kg, whereas our modified platform was stripped down to 2.05 kg. Our modified platform included replacing the standard motor controller with a Pololu 18 volt / 25 ampere simple motor controller \cite{pololu18_25_motor_controller}, and one 5 volt / 5 ampere step-down voltage regulator \cite{pololu5_5_voltage_regulator}. The main benefit of the replaced electronics is that we could power our system with a variety of batteries. Our modified platform was then improved with a constructed frame consisting of 2 x 2 inch by 1/8 inch extruded aluminum plates connecting one 15 inch long section of 80/20 extruded aluminum (1020 Profile) \cite{extruded_aluminum}. Attached to the constructed frame was a Raspberry Pi 3 B+ and an Arduino Mega. The Arduino Mega served as a ROS node for the Raspberry Pi and returned IMU data from an Adafruit LSM9DS1 sensor \cite{adafruit_imu}. The Raspberry Pi was powered by a separate but identical 5 Volt regulator to minimize electrical noise in the power system. Lastly, the Raspberry Pi was also connected to a GlobalSat USB GPS \cite{global_sat_gps}. And although there is a RunCam Camera attached to the 80/20 section, it was unused for this experiment. In total, the platform weighed 3.50 kgs.

The platform communicated to  a WiFi network, although this was only used for SSHing to the platform to execute the ROS launch file. We controlled the platform when necessary with a 4channel TQi controller \cite{tqi_controller}. The extra channels allowed us to deliberately place the platform in autonomous mode or remote controlled mode. In essence this served as a kill switch in case our platform lost control. Our TQi controllers operated in the 2.4 GHz frequency while our WiFi network was switchable between 2.4 GHz and 5 GHz. Our platform was powered by one of two 3S LiPo batteries that delivered over 5400 mAH.

\begin{figure}
\begin{center}
 \vspace{-0.3in}
 \includegraphics[width=4in]{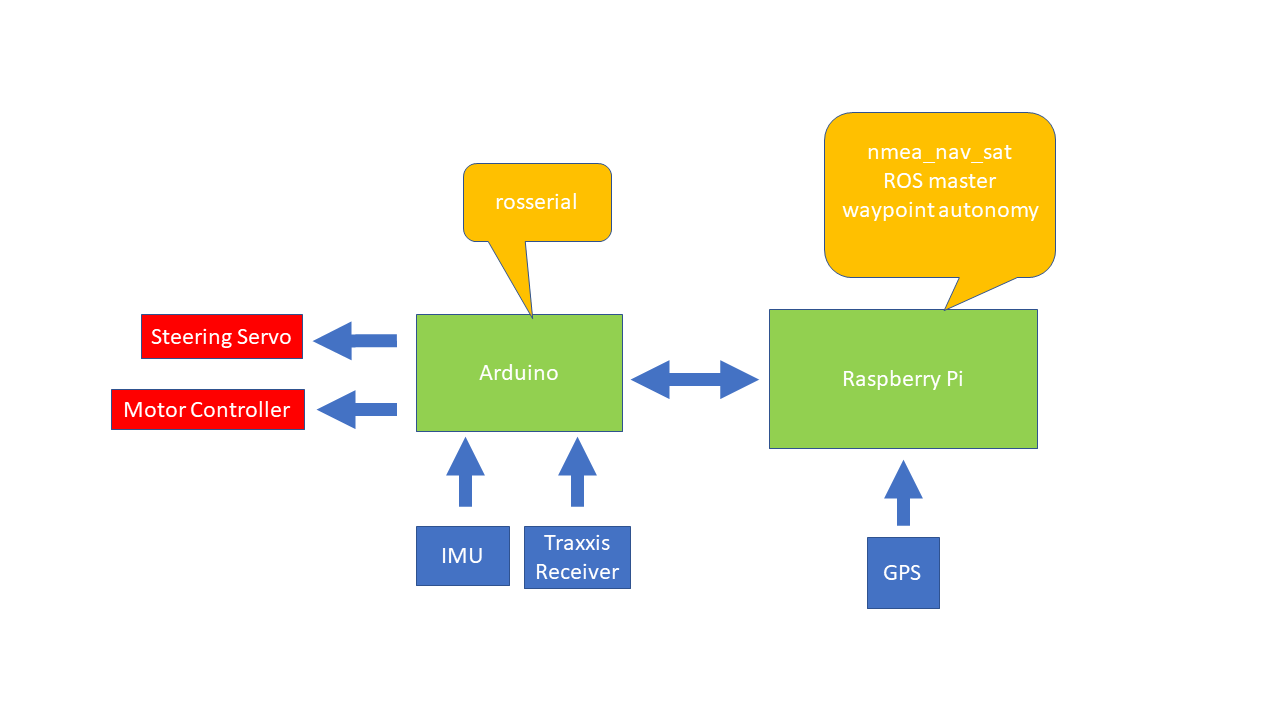}
 \vspace{-0.55in}
 \end{center}
 \caption{UGV hardware and corresponding software diagram.  Actuators are seen in red, sensors are in blue, while compute hardware is in green.  The yellow callout boxes indicate the ROS software nodes corresponding to each compute device.}
 \label{fig:ugv-diagram}
\vspace{-0.2in}
\end{figure}

The hardware and associated software diagram is seen in Figure \ref{fig:ugv-diagram}.
The unmanned ground vehicle leveraged the Robot Operating Software (ROS) as our robot middleware.  The GlobalSat GPS is driven by the NMEA\_NAVSAT\_DRIVER provided by ROS. The UGV is controlled by two inputs.  The arduino receives both inputs and sends messages to the steering and thrust devices.  The first, is our remote control and the second is the Raspberry Pi running ROS.  ROSSerial was used to communicate with the Arduino from the Raspberry Pi.  Through Rosserial, the IMU sent information through the Arduino.  Commands from ROS were then run through Rosserial to the Arduino and out the the steering and thrust devices.  

The ROS mission software developed in simulation was also used for the field trials, with the exception that GPS location and vehicle pitch were read from topics published by the Arduino node. The relationship between the PWM control signals and the mission software  steering and speed signals was established empirically and very coarsely. The lack of careful calibration between simulation and physical platform was purposeful, since closing that gap is part of the objective of the research reported here.

    \section{Results}\label{sec:results}

\begin{figure}
\begin{center}
 \vspace{-0.3in}
(a)\includegraphics[width=3in]{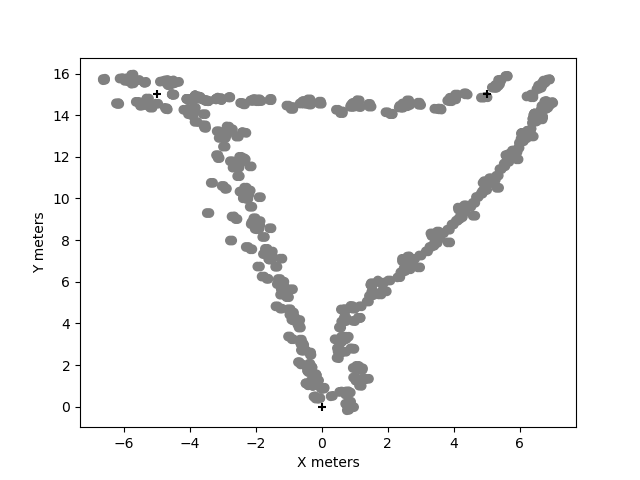}\\
 \vspace{-0.075in}
(b)\includegraphics[width=3in]{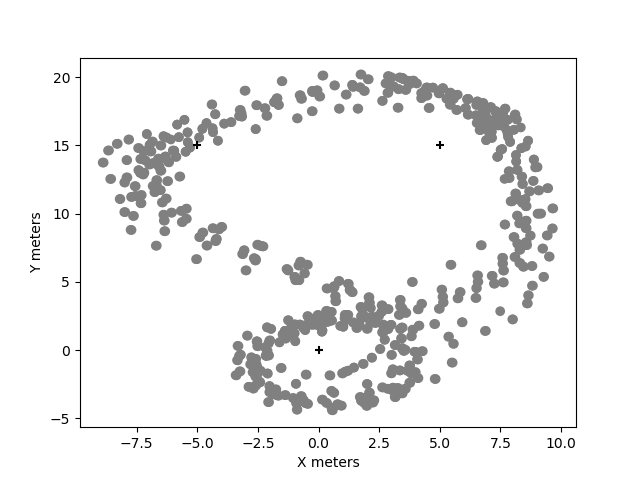}\\
 \vspace{-0.075in}
(c)\includegraphics[width=3in]{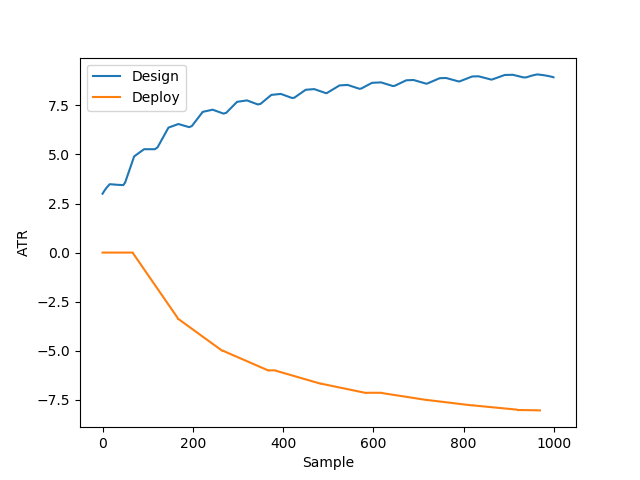}\\
\end{center}
 \caption{Data from design and deployment trials (a) scatter plot of 2D position in design, (b) in deployment, and (c) combined ATR graph. Waypoint locations are marked as `+'.}\label{fig:fieldATR}
  \vspace{-0.2in}
\end{figure}

The data collected from the design and deployment phase is shown in Fig. \ref{fig:fieldATR}. The scatter plots in Fig. \ref{fig:fieldATR}(a) and \ref{fig:fieldATR}(b) show all the 2-D position samples of the simulated robot during design, Fig. \ref{fig:fieldATR}(a), and the physical robot during deployment, Fig. \ref{fig:fieldATR}(b).  The Average Total Reward (ATR) graph shows that the simulation performs well according to its performance monitor expressed as rewards for each iteration that meets the performance requirement. The deployed software fails to meet the performance requirement however, as evidenced by its decreasing ATR graph. We  engineered the field trial to force this effect by introducing a ramp between waypoints (5,15) and (-5,15). 

As described in Algorithm \ref{alg:kc}, the data collected from design and deployment (Eq.(\ref{eqn:h})) was used to generate  the  transition functions $P_{sim}$ and $P_{phy}$ (Eq.(\ref{eqn:sr})). $P_{sim}$ had 951 state transitions, and $P_{phy}$ had 942 transitions. The vast majority of both had a branching factor of 1. There 882 states in common between the two, where states are compared as described in Eq.(\ref{eqn:c}) and its accompanying text. One hundred
paired roll-outs were conducted and divergent states identified. From this, 15 unique divergences were identified and kernels developed for them. Each of the kernels included a linear transfer functions mapping velocity and sensor state to a new sensor state - position, orientation and pitch. 

\begin{figure}
\begin{center}
 \vspace{-0.3in}
(a)\includegraphics[width=3in]{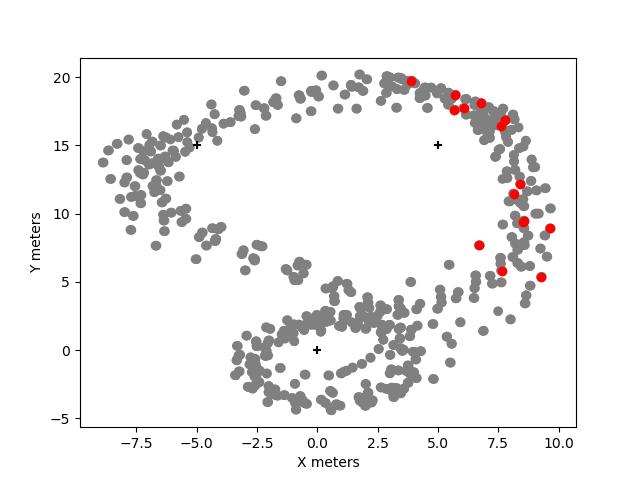}\\
 \vspace{-0.075in}
(b)\includegraphics[width=3in]{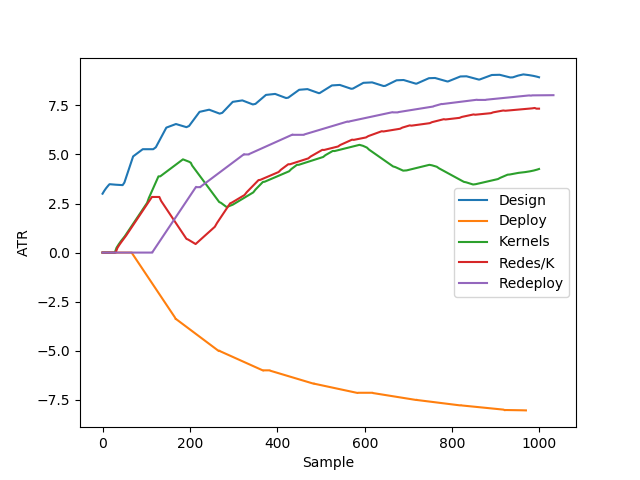}\\
 \vspace{-0.15in}
\end{center}
 \caption{Deployment track data  (a) overlay showing calculated kernel locations in red, (b) ATR graph for kernels, redesign and redeployment. 
 }\label{fig:finalATR}
  \vspace{-0.1in}
\end{figure}

Fig. \ref{fig:finalATR}(a) shows the kernel mean location overlayed on the deployment tracks. Recall that this location is the one immediately preceding the divergence - and on the track overlay, many kernels are clustered around and just after the waypoint preceding the ramp.  

The software was then rerun in the simulation but with the Kernel Manager active. Fig. \ref{fig:finalATR}(b) shows a number of ATR graphs including the original design and deployment graphs. The graph labeled "Kernels" shows the result of executing the original mission software in the kernel modified simulation. The performance is much worse than the original performance, but not as severe as the actual deployment. 

Of course, there are many ways that the simulation could be coerced into failing the performance measure. But unless the mechanism failure allows a designer or learning algorithm to redesign and test in simulation, and that generalizes to improved real performance, the method would have limited use.

In this experiment, the code was manually improved by just adding a conditional statement to check the pitch angle, and to use a greater spring constant and saturation value if the pitch exceeded a threshold. This modification was tested on the kernel modified simulation, producing the ATR graph labelled "Redes/K" in Figure \ref{fig:finalATR}(b). The modified code shows improved performance over the original code, approaching within $\tilde{}25\%$ of the original performance. The same code was redeployed to the robot, and that performance is shown in the ATR graph labeled "Redeploy" in Figure \ref{fig:finalATR}(b). The improvement in behavior transfers from simulation to reality, supporting the argument that policies (software) developed with the kernel mechanism transfer well to reality.

    \section{Conclusions}\label{conclusions}

This paper has proposed a novel approach to the "reality gap" 
problem. This problem arises whether the simulation is being used by human designers or by an automated policy development mechanism, or, as in our case, as part of an approach to self-healing autonomous system software \cite{Lyons_2020}. 
We propose an approach that collects simulation and real world observations and builds conditional probability functions for them. These are used to generate paired roll-outs and look for points of divergence in behavior. These are used to generate state-space kernels  to coerce the simulation into behaving more like observed reality within that region of the state space. 

The method was evaluated in field trials using ROS/Gazebo for simulation and a heavily modified Traaxas platform for outdoor deployment. Our results support not just that the kernel approach can force the simulation to behave more like reality, but that the modification is such that an improved control policy tested in the modified simulation also performs better in the real world.

The kernel managed simulation in our field trials did not produce as poor an ATR graph as the actual deployment Fig. \ref{fig:finalATR}. This is principally due to restricting the scope of each state-space kernel to a small region of state space with $e_k \circ c_k(s') < \epsilon_c.$ The advantage is that it allows us to leverage a fast function approximation method. The disadvantage is that it could require making a lot of kernels to capture the real environment with high fidelity. We argue however that there is an advantage in understating the effect of experience gained in each iterative deployment since it produces a sequence of smaller learning obstacles rather than a single large obstacle. This has been shown to be a useful learning strategy \cite{Wang_2019}.

The work most similar to ours is that of \cite{Chebotar_2019}. A crucial point of difference is that \cite{Chebotar_2019} (and some others) address the reality gap problem with a principled domain randomization approach, providing a range of environment for policy development, and generating a policy that it is robust along the right dimensions of variability. We address the same problem but from the perspective of making each simulation run more closely resemble reality. This is reflected in how the simulation is 'wrapped' by each approach: our approach requires a more invasive configuration -- access to the Gazebo model information -- but we argue that our configuration is less application specific.





\end{document}